\documentclass{article} 
\usepackage{iclr2022_conference,times}

\usepackage{amsmath,amsfonts,bm}

\def\eqref#1{equation~\ref{#1}}

\def\1{\bm{1}}

\DeclareMathAlphabet{\mathsfit}{\encodingdefault}{\sfdefault}{m}{sl}
\SetMathAlphabet{\mathsfit}{bold}{\encodingdefault}{\sfdefault}{bx}{n}

\usepackage{hyperref}
\usepackage{url}

\usepackage{graphicx}
\usepackage{multirow}
\usepackage{amsmath}
\usepackage{booktabs}

\title{Scaling Laws for the Few-Shot Adaptation of Pre-trained Image Classifiers}

\author{Gabriele Prato\textsuperscript{1,2}, Simon Guiroy\textsuperscript{1,2}\thanks{Equal contribution.} , Ethan Caballero\textsuperscript{1,2}\footnotemark[1] , Irina Rish\textsuperscript{1,2}, Sarath Chandar\textsuperscript{1,3} \\
\textsuperscript{1}Mila \\
\textsuperscript{2}Université de Montréal \\
\textsuperscript{3}École Polytechnique de Montréal \\
\texttt{pratogab@mila.quebec}}

\iclrfinalcopy 
\begin{document}

\maketitle

\begin{abstract}
Empirical science of neural scaling laws is a rapidly growing area of significant importance to the future of machine learning, particularly in the light of recent breakthroughs achieved by large-scale pre-trained models such as GPT-3, CLIP and DALL-e. Accurately predicting the neural network performance with increasing resources such as data, compute and model size provides a more comprehensive evaluation of different approaches across multiple scales, as opposed to traditional point-wise comparisons of fixed-size models on fixed-size benchmarks, and, most importantly, allows for focus on the best-scaling, and thus most promising in the future, approaches. In this work, we consider a challenging problem of few-shot learning in image classification, especially when the target data distribution in the few-shot phase is different from the source, training, data distribution, in a sense that it includes new image classes not encountered during training. Our current main goal is to investigate how the amount of pre-training data affects the few-shot generalization performance of standard image classifiers. Our key observations are that (1) such performance improvements are well-approximated by power laws (linear log-log plots) as the training set size increases, (2) this applies to both cases of target data coming from either the same or from a different domain (i.e., new classes) as the training data, and (3) few-shot performance on new classes converges at a faster rate than the standard classification performance on previously seen classes. Our findings shed new light on the relationship between scale and generalization.

\end{abstract}

\section{Introduction}
Over the past decade, deep learning has made tremendous progress in multiple fields, especially  in vision \citep{alam2020survey} and  natural language processing \citep{2020arXiv200301200T}. However, several important issues remain unsolved, including  the ability to generalize well to novel, out-of-distribution  data \citep{2021arXiv210302667A}. A particularly challenging situation involves  simultaneous changes at test time  in both the input  and the task, class distributions, $p(x)$ and  $p(y|x)$. For example, a self-driving car seeing an elephant for the first time should be able to recognize it as a "new  object", while  seeing another elephant afterwards, it should be able to recognize it as the {\em same} "new object".  Obviously, any  deployment  of deep networks in the real world will likely require  them to deal with new situations not encountered during training. Recent findings \citep{2020arXiv200514165B,2021arXiv210300020R,2021arXiv210212092R} suggest that pretraining large scale models on large scale datasets may improve their generalization to novel data and tasks. 

The Deep Learning book \citep{Goodfellow-et-al-2016} states that: "for nonparametric models, more data yield better generalization until the best possible error is achieved." While neural networks are usually parametric models, practitioners tend to adjust their capacity for a given problem based on performance. This is a form of nonparametric learning algorithm. The statement holds because of the i.i.d. assumption, but what happens when samples from the train and test set are not identically distributed? Can more training samples still help in this case? According to the no free lunch theorem \citep{wolpert1997no}, not if we consider all possible data-generating distributions. In practice though, assumptions can be made about the kind of probability distributions that we encounter, since we are interested in real world problems. So then what happens in such cases, does more training data help?

One real world domain where assumptions can be made about the probability distribution is that of vision. Scaling on real world vision tasks can potentially improve generalization on other real world vision problems. Because of its widespread use, vision is a very interesting case study. It's a simple, yet very important problem and a very good example of real world application. Indeed, it is hard to imagine neural networks being deployed in the real world without vision. Such agents will quite certainly encounter situations not seen during training. Knowing how these models will perform in these situations is very important.

In this work, we empirically study the effect of scaling the training set size as well as the number of classes on the generalization error. Specifically, we are interested in the effect of scale in the vision domain. For our study, we train standard image classifiers on various datasets for multiple different values of our scaling parameters. We then measure their few-shot performance on new classes using multiple datasets and evaluation methods. Few-shot performance is measured via the fine-tuning of the classification layer, all other weights remaining frozen, as well as the matching network \citep{2016arXiv160604080V} and prototypical network \citep{2017arXiv170305175S} method. We find the effect of scale on performance to follow the same trend for all three methods. Our analysis is performed on widely used architectures in the field of vision, trained and tested on ten datasets, ranging from natural to non-natural images.

By performing this study, our goal is to shed some light on the effect of scale on the generalization performance.  Neural scaling laws research emerged quite recently, triggered by successes in large-scale pretrained models, as it aims to better understand promises and limitations of scaling. The importance of approaching learning methods from an empirical science perspective and  discovering scaling laws with respect to various factors, including, but not limited to, the data and model size, is being more and more widely recognized in the deep learning community. Indeed, it provides multiple advantages including a more rigorous and comprehensive methodology for comparing competing deep learning approaches, and choosing the best-scaling ones as they are most likely to withstand the test-of-time challenge.  

\begin{figure}
  \begin{center}
  \includegraphics[width=1.0\linewidth]{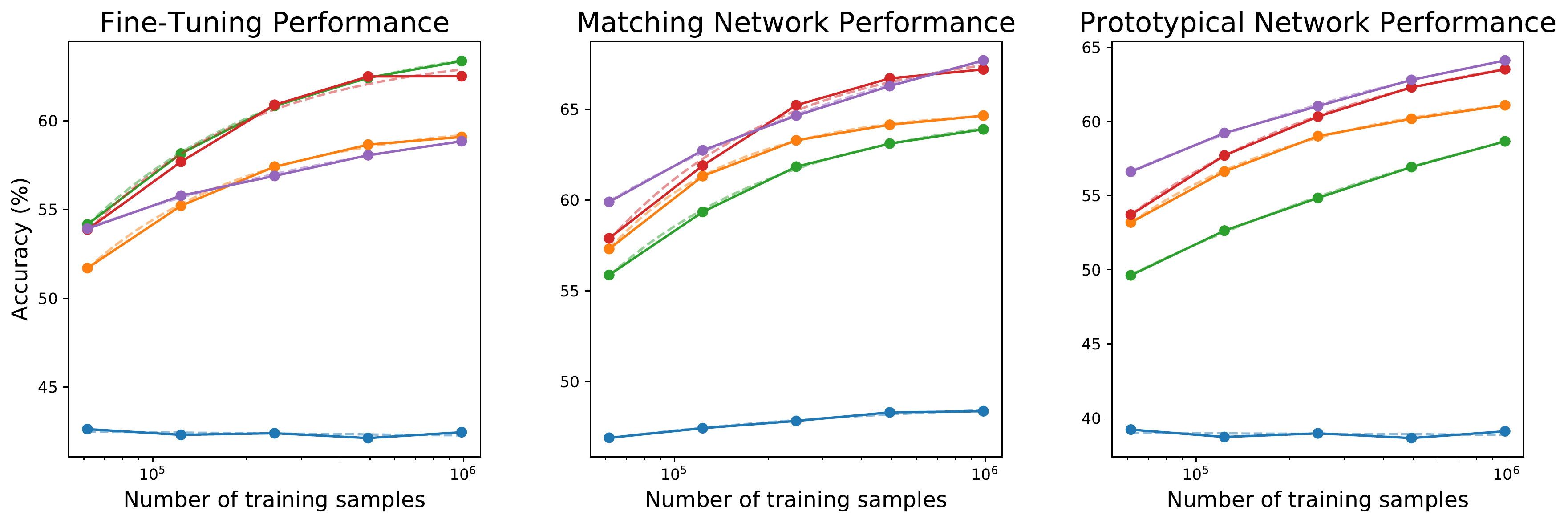}
  \includegraphics[width=0.65\linewidth]{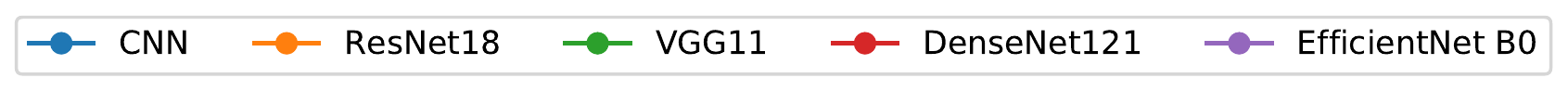}
  \includegraphics[width=0.35\linewidth]{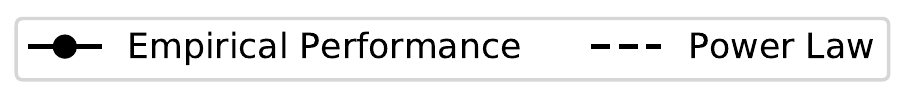}
  \end{center}
  \caption{ImageNet one-shot performance averaged over all target datasets: Aircraft, Bird, COCO, Describable Texture, Flower, Fungi, Omniglot, Quickdraw and Traffic Sign. (Left) Fine-tuning performance, (Center) Matching Network performance and (Right) Prototypical Network performance. The one-shot performance scales with the training set size following simple power laws.}
  \label{fig:results_summary}
\end{figure}

Our key observations are the following:
\begin{itemize}
    \item We find that few-shot performance improves on average as training data scales. Even for multiple distant train-target pairs such as Fungi to Aircraft, we find that scaling the train dataset improves the few-shot performance on the target dataset.
    \item Just like standard classification performance improvements, few-shot performance improvements can be well-approximated by power laws, as a function of training set size.
    \item We find that when scaling the training set size, the few-shot performance on new classes converges at a faster rate than the standard test classification performance on classes seen during training. This highlights the importance of studying scaling laws in more settings than simply the standard analysis of scaling effect on in-distribution performance.
\end{itemize}

\section{Experimental Setup} \label{sec:empirical_setup}
In this section, we detail the training and evaluation procedure common to all experiments as well as the power laws we fit.

\subsection{Training \& Evaluation Procedure}
In all cases, models are trained as standard image classifiers and evaluated via fine-tuning of the classification layer as well as matching networks and prototypical networks. Best few-shot performance is reported for each model, measured throughout training. See the following section for a detailed explanation of the fine-tuning process, matching networks and prototypical networks.

The ten datasets from Meta-Dataset \citep{2019arXiv190303096T} are used in our study: Aircraft, Bird, COCO, Describable Texture, Flower, Fungi, ImageNet, Omniglot, Quickdraw and Traffic Sign. 80\% of each dataset's training classes are used as the training set, the other 20\% is kept for few-shot evaluation when the target dataset is the same as the train dataset, e.g. Aircraft-Aircraft. When the target dataset is different than the train dataset, we evaluate on all classes and all data of that target dataset.

We perform 5-way 1-shot and 5-way 5-shot evaluation, repeated over thousands of trials. For each trial, a set of five classes is chosen at random and one sample is drawn for each class in the case of 1-shot and five in the case of 5-shot. These samples are used as the support set. Then, samples are drawn at random for each of the five classes and classified using the fine-tuning, prototypical network and matching network method.

Any additional detail is provided in the Appendix section \ref{sec:appendix_empirical_setup}.


\subsection{Fine-Tuning}
For each trial, the fine-tuning evaluation method consists of training a new classifier for the given set of five classes. The original classification layer is swapped with a newly randomized one at each trial and is trained on the support set. Only the classification layer is trained, the remaining pre-trained weights remain frozen. The support set, five samples in the case of one-shot, twenty-five in the case of 5-shot, forms a batch with which the model classifies five times, each time taking a gradient step. The fine-tuned model is then used to classify non-support set samples for each of the five classes.

\subsection{Prototypical Networks}
Instead of outputting a logit for each class, prototypical networks \citep{2017arXiv170305175S} compute a metric between the model's output and each class prototype. The result is then used to predict to which class the input sample belongs to. Typically, as is our case, the classification layer is discarded and the final latent representation is used as the output instead. Prototypes are computed by averaging the output embedding of same class samples from a support set. Common metric choices are euclidean distance and cosine similarity, although any function could be used. We use euclidean distance in our study.

The advantage of prototypical networks is that they are not constrained to making predictions for samples only belonging to the training classes. Prototypes can be computed with samples from classes never seen during training.

\subsection{Matching Networks}
In the case of matching networks \citep{2016arXiv160604080V}, we also discard the classification layer and use the penultimate latent representation instead to perform classification. To classify a sample, the cosine similarity is measured between its encoded representation and the one of each other sample in the support set. Then these cosine similarities are softmaxed and the resulting probabilities are summed by class. The class with the highest probability is chosen as the predicted label. In the one-shot case, matching networks and prototypical networks are equivalent, except that the distance is measured via cosine similarity in the case of matching networks and euclidean distance in the case of prototypical networks.

\subsection{Power laws}
For each train-target pair and each model, we fit when possible simple power laws for both the scaling of the training set size $N$ and the number of training classes $C$:
\begin{align}
    Err(N) &= Err_{\infty} + k N^\alpha, \\
    Err(C) &= Err_{\infty} + k C^\alpha,
\end{align}
where $Err$ is the estimated error rate, $Err_{\infty}$ the irreducible error rate and $kN^\alpha$ is the reducible error rate with model-dependent scaling constants $k$ and $\alpha$.

\begin{table}
  \caption{Power law constants for ImageNet one-shot performance averaged over all target datasets. See Figure \ref{fig:results_summary}. $N$ is the training set size and $Err(N)$ is the error rate.}
  \label{tab:imagenet_average_power_law_constants}
  \begin{center}
  \resizebox{\linewidth}{!}{
  \begin{tabular}{lccc}
    \toprule
    Model & Fine-Tuning $Err(N)$ & Matching Network $Err(N)$ & Prototypical Network $Err(N)$ \\
    \midrule
    ResNet18 & $39.95 + \left(\frac{N}{8.18 \times 10^5}\right)^{-0.82}$ & $34.95 + \left(\frac{N}{4.25 \times 10^5}\right)^{-1.06}$ & $37.55 + \left(\frac{N}{1.54 \times 10^6}\right)^{-0.69}$ \\
    VGG11 & $34.89 + \left(\frac{N}{2.19 \times 10^6}\right)^{-0.67}$ & $34.68 + \left(\frac{N}{1.54 \times 10^6}\right)^{-0.70}$ & $32.51 + \left(\frac{N}{5.59 \times 10^9}\right)^{-0.25}$ \\
    DenseNet121 & $36.02 + \left(\frac{N}{1.11 \times 10^6}\right)^{-0.81}$ & $31.22 + \left(\frac{N}{1.46 \times 10^6}\right)^{-0.76}$ & $33.65 + \left(\frac{N}{6.62 \times 10^6}\right)^{-0.54}$ \\
    EfficientNet B0 & $38.41 + \left(\frac{N}{1.54 \times 10^7}\right)^{-0.37}$ & $27.72 + \left(\frac{N}{7.68 \times 10^7}\right)^{-0.35}$ & $30.10 + \left(\frac{N}{3.51 \times 10^8}\right)^{-0.30}$ \\
    \bottomrule
  \end{tabular}
  }
  \end{center}
\end{table}

\section{Empirical Results} \label{sec:empirical_results}
The following sections detail our findings for each scaled parameter. Full results are available in the Appendix section \ref{sec:appendix_full_results}.

\subsection{Scaling Amount of Training Data} \label{sec:scaling_training_data}
The first parameter that we scale is the amount of training data. Five ratios are used: 100\%, 50\%, 25\%, 12.5\% and 6.25\% of the total training data for each dataset. The amount of training classes remains the same no matter the ratio. Because of Flower's dataset size, we only trained with 100\%, 50\%, 25\% and 12.5\% of the data. For each ratio, we train five architectures: a four-layer convolutional network \citep{LeCun:1989:BAH:1351079.1351090}, a VGG11 \citep{2014arXiv1409.1556S}, a ResNet18 \citep{2015arXiv151203385H}, a DenseNet121 \citep{2016arXiv160806993H} and an EfficientNet B0 \citep{2019arXiv190511946T}. All five models are trained and tested on all datasets.

On average, when scaling the amount of training data, we find that the out-of-distribution performance of the image classifiers improves. See Figure \ref{fig:results_summary} for the average few-shot performance of models trained on ImageNet. Results for other train datasets can be found in the Appendix \ref{sec:appendix_full_results}. The only model which does not improve on average is the CNN. Only with matching networks did the performance improve. We believe this might be due to the CNN's shallow architecture and its classification layer representing most of the weights of the model. At test time, since we discard this classification layer, we discard most of the weights of the model. The latent representation we get from the resulting shallow network does not seem to benefit from more training data.

\begin{figure}
  \begin{center}
  \includegraphics[width=1.0\linewidth]{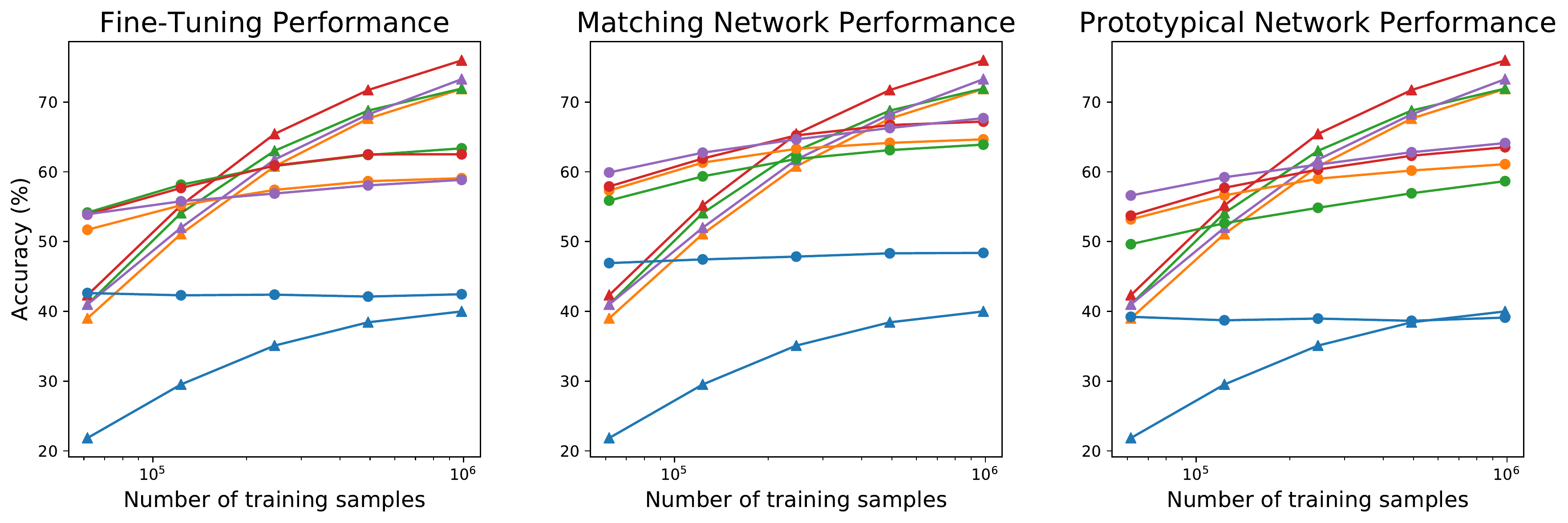}
  \includegraphics[width=0.65\linewidth]{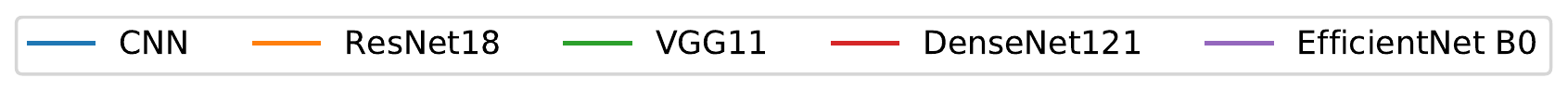}
  \includegraphics[width=0.5\linewidth]{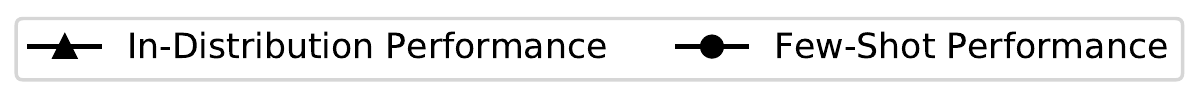}
  \end{center}
  \caption{Comparison of the standard classification performance on classes seen during training (in-distribution performance) and the few-shot performance. Standard classification performance is on the ImageNet test set for classes seen during training. Few-shot performance is the one-shot performance averaged over all target datasets: Aircraft Bird, COCO, Describable Texture, Flower, Fungi, Omniglot, Quickdraw and Traffic Sign.}
  \label{fig:results_summary_with_powerlaws}
\end{figure}

\begin{figure}
  \begin{center}
  \includegraphics[width=1.0\linewidth]{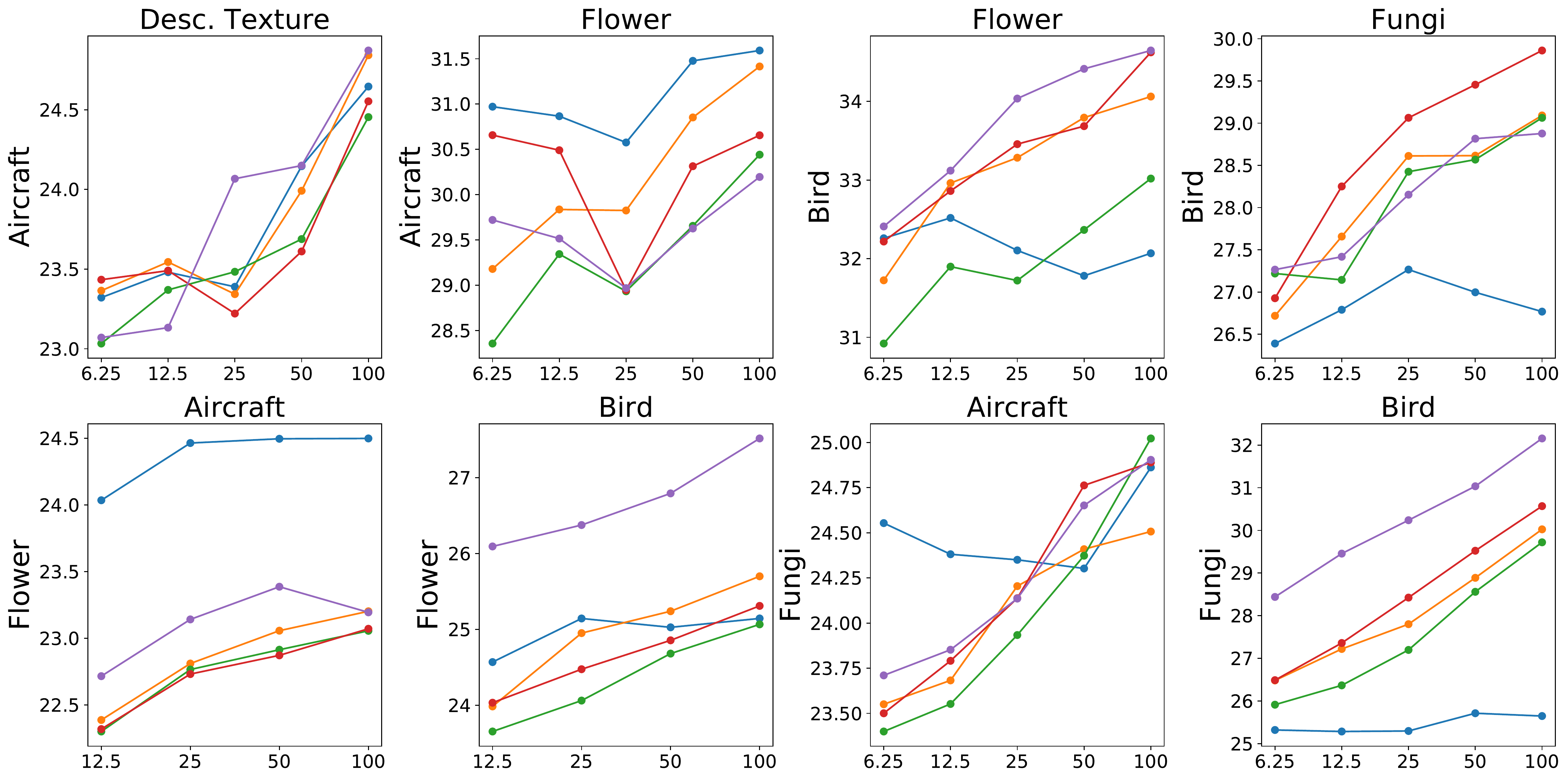}
  \includegraphics[width=0.5\linewidth]{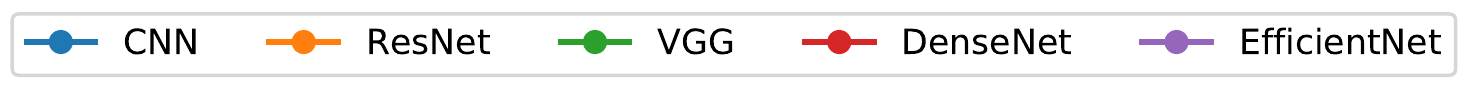}
  \end{center}
  \caption{Few-shot performance for multiple train-target pairs of natural image datasets. For each plot, the train dataset is written on the left and the target dataset on top. X-axis is the percentage of the total training data and y-axis is the 5-way 5-shot accuracy. For 5-way 1-shot accuracy, see section \ref{sec:appendix_full_results}. Both 5-way 5-shot and 1-shot follow similar trends.}
  \label{fig:distant_pairs}
\end{figure}

As a comparison with the few-shot performance, we looked at how scale had an effect on the standard classification test performance on classes seen during training. What we found is that the few-shot performance converges at a faster rate than the standard classification performance, see Figure \ref{fig:results_summary_with_powerlaws}. This is important, as it will be much more more costly to improve the few-shot performance via scaling the training set size than the standard classification performance.

On average, when the training dataset is one of natural images (Aircraft, Bird, COCO, Describable Texture, Flower, Fungi and ImageNet), the few-shot performance on other natural image datasets does seem to improve, see Figure \ref{fig:distant_pairs}. We find this very interesting, as it is not so obvious if training on more fungi for example can make a model better at classifying aircrafts. This suggest that there are features even in specific natural domain which generalize well to all sorts of other natural domains.

\begin{figure}
  \begin{center}
  \includegraphics[width=1.0\linewidth]{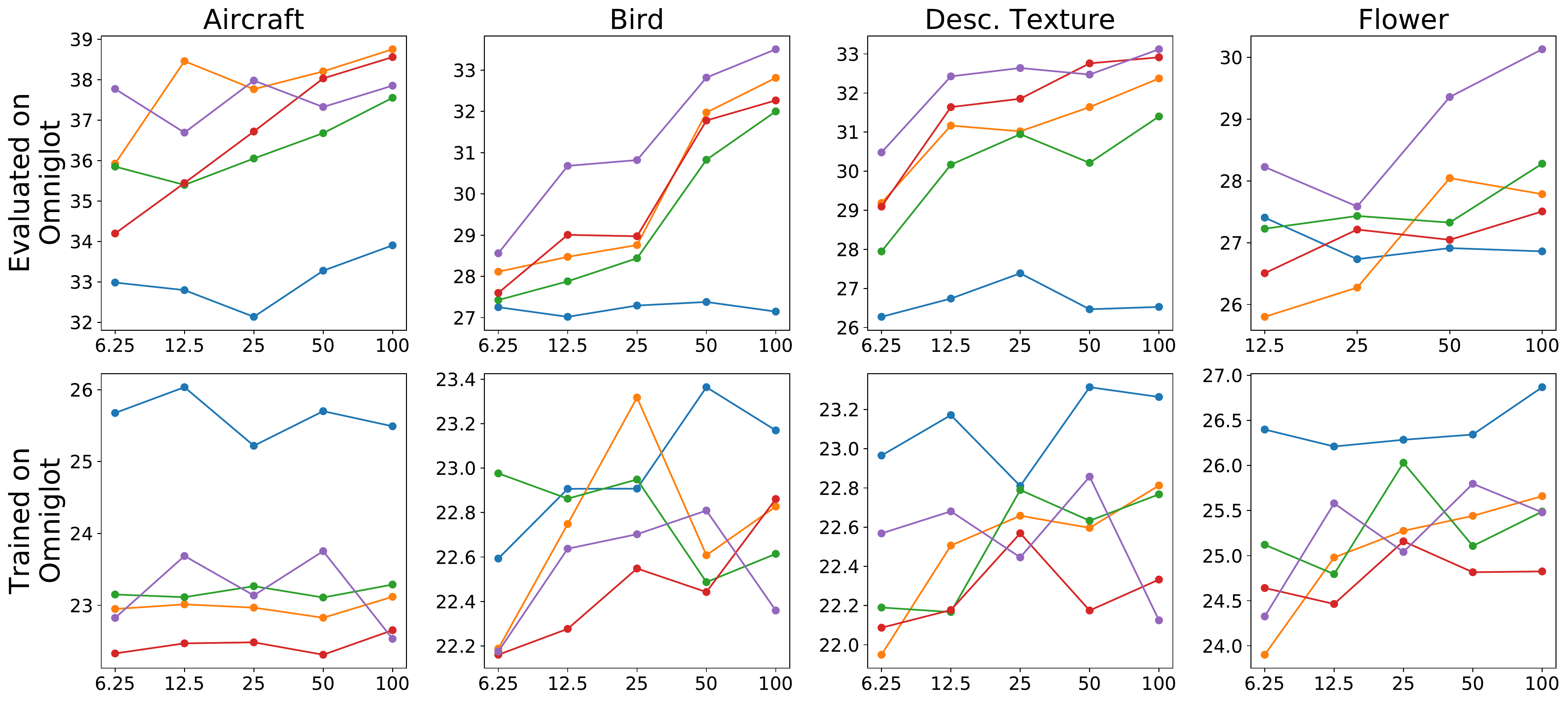}
  \includegraphics[width=0.5\linewidth]{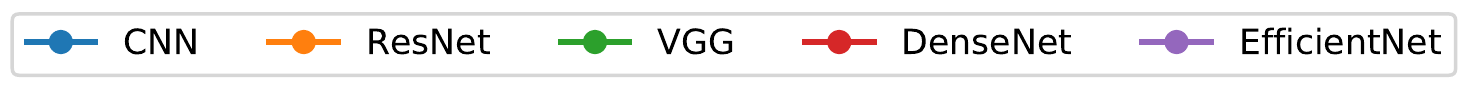}
  \end{center}
  \caption{(Top row) models trained on the dataset marked on top of each plot and evaluated on Omniglot. (Bottom row) models trained on Omniglot and evaluated on the dataset marked on top of each respective column. X-axis is the percentage of the total training data and y-axis is the 5-way 5-shot accuracy. For 5-way 1-shot accuracy, see section \ref{sec:appendix_full_results}. Both 5-way 5-shot and 1-shot follow similar trends.}
  \label{fig:omniglot_results}
\end{figure}

As for few-shot performance on non-natural image datasets, surprinsingly, scaling the training datasets such as Aircraft, Bird, Describable Texture, Flower and Fungi results in better performance on Omniglot, a character dataset, see Figure \ref{fig:omniglot_results}. This shows that training on bigger natural image datasets can also potentially improve the generalization on non-natural image datasets. This also highlights the unintuitivity in predicting whether or not training on more data from a certain domain will result in better performance on some other domain. Indeed, in both the natural and non-natural cases, we expected that for the few-shot performance to improve, distributions would have to be much closer. Inversely though, training on more Omniglot data does not seem on average to provide as much gains on other datasets, see Figure \ref{fig:omniglot_results}. This suggests that while features in the natural domain can be useful for non-natural domains, the inverse is not.


Finally, the only training dataset where on average scaling does not seem to improve the few-shot performance is the Traffic Sign dataset. As an evaluation dataset, apart from a few exceptions, few-shot performance on Traffic Sign does not seem to improve much as training data is scaled. We believe these issues might be due to the nature of the task, where many classes share key features as well as potentially the low quality of the data. See section \ref{sec:appendix_full_results} for results.

\subsection{Scaling Amount of Training Classes} \label{sec:scaling_training_classes}
\begin{figure}
  \begin{center}
  \includegraphics[width=1.0\linewidth]{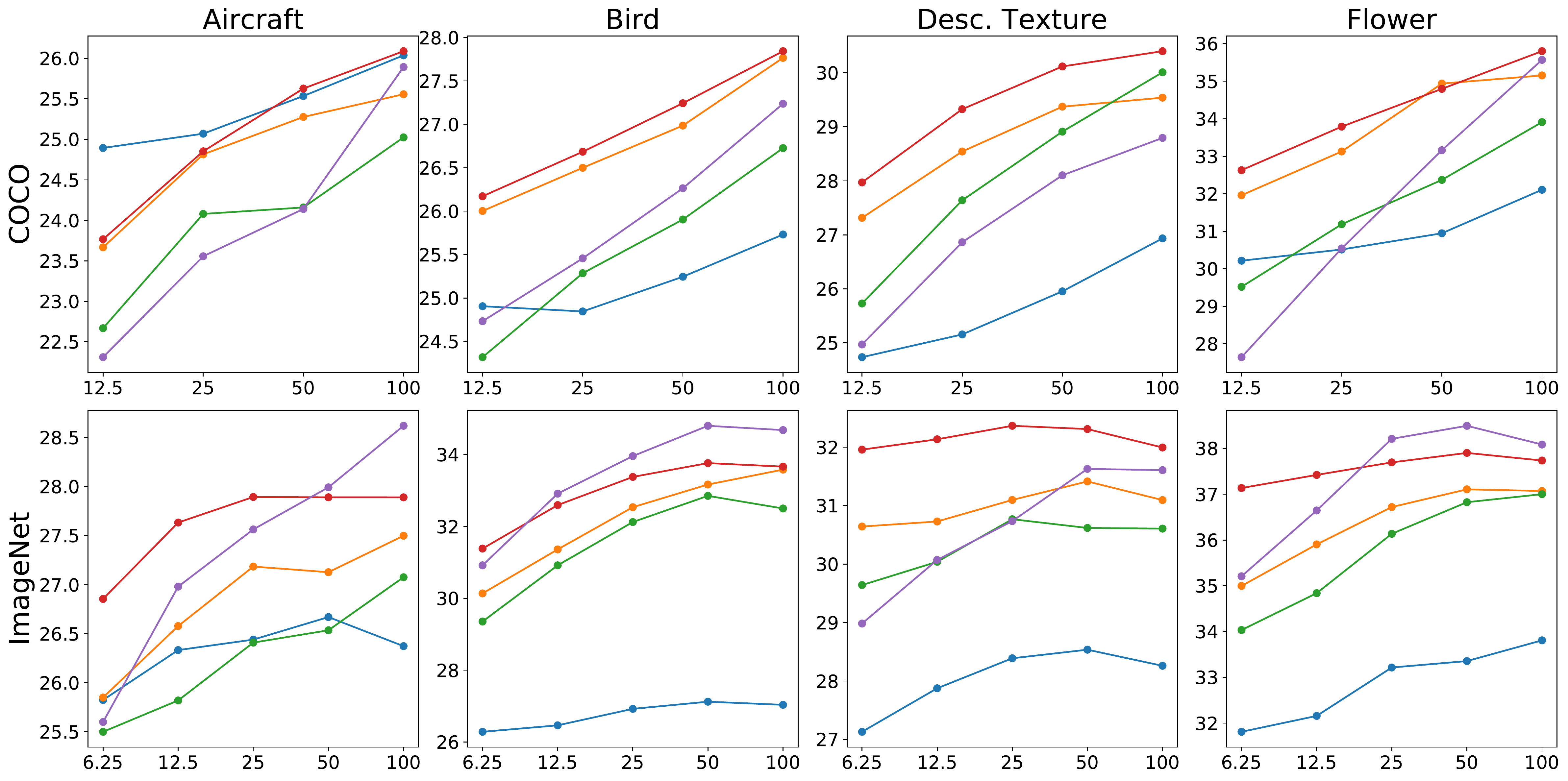}
  \includegraphics[width=0.625\linewidth]{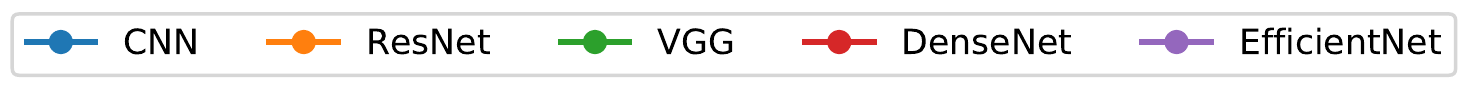}
  \end{center}
  \caption{Scaling number of training classes results for various train-target pairs. For each plot, models are trained on the dataset marked on the left of each respective row and evaluated on the dataset marked top of each respective column. X-axis is the percentage of the total number of training classes of each dataset and y-axis is the 5-way 5-shot accuracy. For 5-way 1-shot results, see section \ref{sec:appendix_full_results}. Both 5-way 5-shot and 1-shot follow similar trends.}
  \label{fig:scaling_amount_training_classes}
\end{figure}

In the last section, the number of training classes was fixed for all training data ratios. We now take a look at the inverse setup, where the amount of training classes scales, while the amount of training data remains the same. In this setup, we are interested in seeing if more classes helps with generalization.
We use four ratios for our experiments: 100\%, 50\%, 25\% and 12.5\%. 6.25\% is left out due to the low number of classes in most datasets. Again, we train the same five architectures on all ten datasets for all ratios and evaluate on the same seven datasets as in section \ref{sec:scaling_training_classes}.

On average, see Figure \ref{fig:results_summary} and \ref{fig:results_summary_with_powerlaws}, when scaling the amount of training classes, few-shot performance seems to improve at lower percentages, but then plateaus or start to plateau at some point, the only exception being the CNN. Estimated power laws in Figure \ref{fig:results_summary_with_powerlaws} for the ResNet and DenseNet models would potentially be better with more data points to fit the function.


Omniglot, Fungi and ImageNet are the datasets with the most number of train classes, 1296, 1120 and 800 respectively. Scaling the amount of training classes for Fungi and Omniglot does not seem on average to improve the few-shot performance. For ImageNet though, more classes does seem to help, but the gains seem to level off after more than 200 classes. See Figure \ref{fig:scaling_amount_training_classes}.

COCO and ImageNet are the datasets with the most diverse set of classes. While gains when scaling on ImageNet seem to flatten towards the end, for COCO scaling seems to improve the few-shot performance on every target dataset, see Figure \ref{fig:scaling_amount_training_classes}. COCO only has a total of 64 train classes though, which could explain why the performance keeps improving. When the number of classes is low in the ImageNet experiments, we also see steady gains on average.

For other datasets, we find results to be mixed. In general though, it seems like when the number of classes is low, more is better, but up to a certain point. It also seems like more from a diverse set of classes helps more than more classes from a more specific domain. 

\section{Related Work}
Recently there has been a gain in interest in studying the effects of scale in machine learning, although only few studies have been done so far. \citet{2017arXiv171200409H} are the first to study the manner in which deep learning scales predictably empirically via a power law functional form. They specifically study how loss scales with respect to training dataset size on language modelling, image classification, and speech recognition tasks.

\citet{2018arXiv181103259Z} perform multiple experiments where they isolate features such as number of element and color and check if GANs \citep{2014arXiv1406.2661G} and VAEs \citep{2013arXiv1312.6114K} manage to generalize or not. For example, when training with images containing always three objects, they show that generative models do generate samples containing fewer, equal and more objects. They show the same thing happening for color and when mixing these features. They find that when the number of combinations is low in the training set, the model can simply learn to memorize them. When they increase the number of combinations though, that's when the model starts to generate new combinations.

More recently, \citet{2020arXiv200108361K} study scaling laws for Transformer \citep{2017arXiv170603762V} language models. They find that performance has a power law relationship with the dataset size, model size and amount of training individually when not bottlenecked by the other two parameters. Additionally, for an equal number of parameters, tuning within a reasonable range the Transformer's depth versus width has little impact on performance. They also demonstrate that overfitting follows a simple ratio between model and dataset size. They show that larger models are more sample efficient and that when training compute is on a budget, but model and dataset size is not, optimal performance is obtained by training very large models and early stopping, rather than training smaller models to convergence. They also find that, when out-of-distribution, the language modelling performance correlates with the validation performance of the training problem, offset by a constant.

Similarly, \citet{2020arXiv200514165B} train language models of various sizes, then evaluate them on multiple downstream natural language processing tasks without any fine-tuning and restrictions on the amount of training data or training compute budget. They show that performance scales in relationship with model size.

As for CNNs, \citet{2020arXiv200708558D} evaluate the impact of the model size and dataset size on robustness, where classes at train and test time are the same, but there is a distribution shift in the data, for example changes in the lighting of image samples. They find that scaling both model size and training set size improves such robustness.

Other work studying the impact of scale on Transformers, \citet{2020arXiv201014701H} train autoregressive Transformers on tasks from different domains, including vision, language and math. The latter is the only domain for which they evaluated outside of the training distribution. Models were evaluated on increasingly difficult problems beyond the difficulty of the training set. In all cases they find that loss scales with respect to model size and amount of training.

\citet{2021arXiv210300020R} train a model to match images with their captions. They vary the amount of compute and show that zero-shot error averaged over multiple vision datasets scales as a function of compute.

\citet{2021arXiv210201293H} use language models pretrained on text and fine-tuned on python code to show that when the fine-tuning dataset is small, models trained from scratch eventually hit a performance wall no matter the model size, while equivalently sized pre-trained models keep improving. They also show that the smaller the fine-tune dataset, the more larger models help when pre-trained compared to training from scratch. To reach a certain loss, pre-trained models need far less training examples than models trained from scratch. This difference scales with respect to model size and inversely with the size of the fine-tuning dataset.

\citet{2021arXiv210305247L} pretrain a Transformer with a language modelling task. Then they freeze the feed-forward and self-attention weight and finetune the rest of the model on non-language tasks such as image classification, protein folding and numerical computation. They show that even with such a drastic change in modality, the models with frozen weights perform just as well as Transformers trained from scratch on each individual task.

\section{Discussion}\label{sec:limitations}
The initial motivation for this work was to study the effect of scale on the out-of-distribution generalization of neural networks. Since there exists no method to evaluate the classification performance of neural networks on new classes without requiring at least some learning, we decided to analyze the impact of scale on the few-shot performance instead. Considering that all the weights in our study remain frozen at test time, except for the final classification layer in the fine-tuning case, we believe that this provides some idea of how scale impacts the out-of-distribution performance of neural networks. We thus expect the out-of-distribution performance to converge at a faster rate than the in-distribution performance. Additional work needs to be done in order to better understand the relationship between scale and out-of-distribution generalization.

Out-of-distribution generalization is a major problem in machine learning and studying its relationship with scale is very important. Many problems are considered out-of-distribution generalization problems and we believe it is important to understand which types of problems will be most impacted by scale and which won't such that research can be better focused on solving issues not mitigated by scale.



\section{Conclusion}
In this work, we studied the effect of scale on the few-shot performance of image classifiers. We found that in general, scaling the training set size improves few-shot performance and can be well-approximated via power laws. We also found the few-shot performance on new classes to converge at a faster rate than the standard classification performance on classes seen during training. This highlights the importance of studying scaling laws in more than just the standard analysis on the effect of scale on the in-distribution performance.

The question of whether or not scale improves generalization in the few-shot learning scenario is an important question and part of a broader scope trying to understand the effect of scale on generalization. We hope that our work will inspire more future work in this direction as well as provide useful insights to practitioners.

\bibliography{references}
\bibliographystyle{iclr2022_conference}

\appendix
\section{Appendix}
\subsection{Empirical Setup Additional Details} \label{sec:appendix_empirical_setup}

\subsection{Full Empirical Results} \label{sec:appendix_full_results}

\begin{figure}
  \begin{center}
  \includegraphics[width=0.87\linewidth]{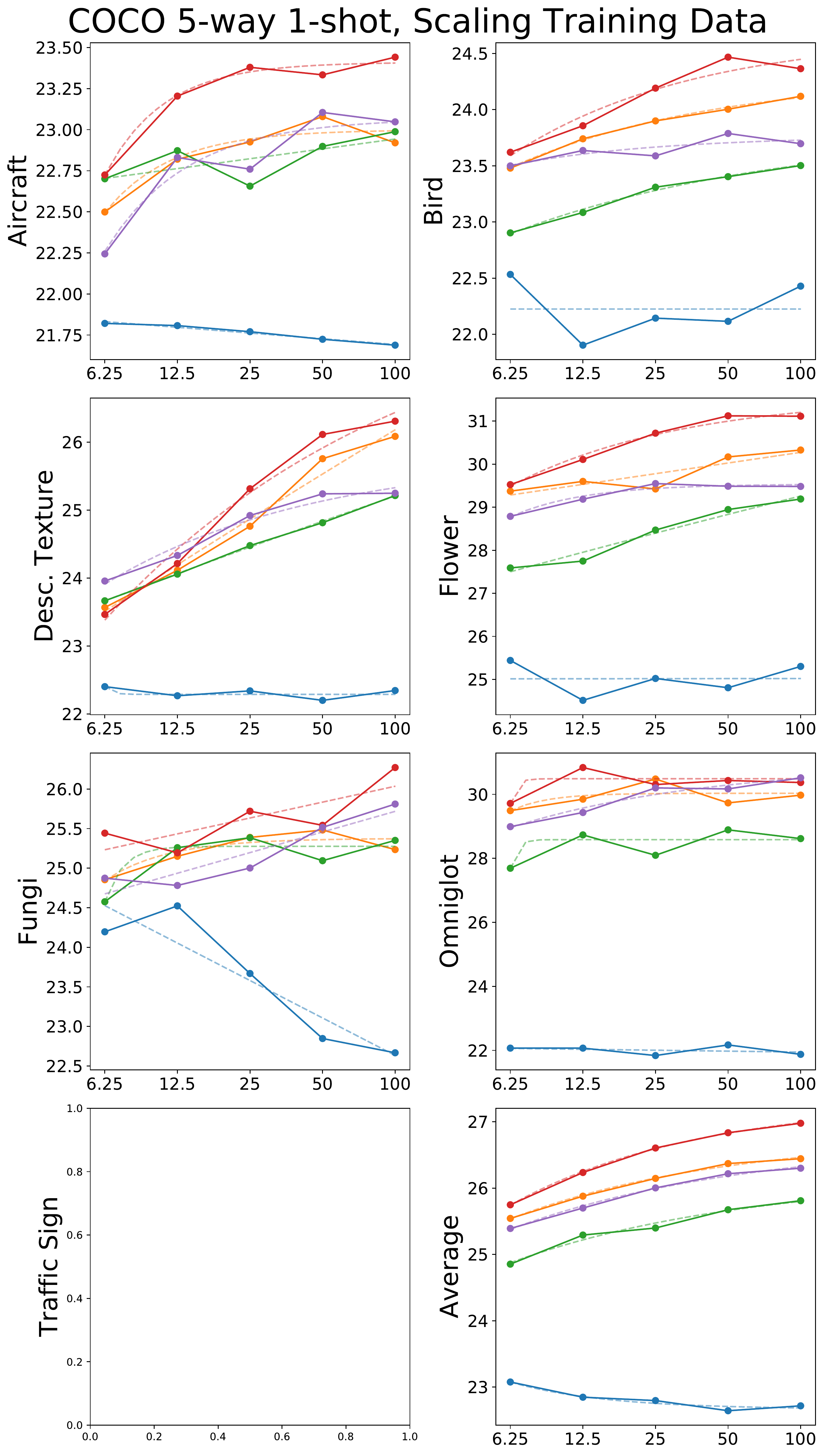}
  \includegraphics[width=0.55\linewidth]{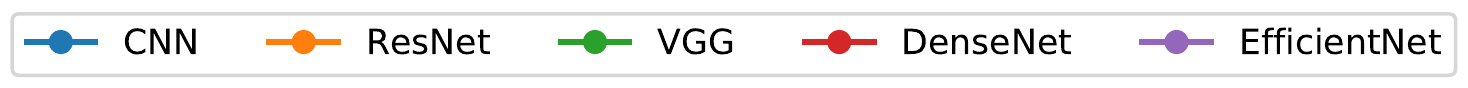}
  \end{center}
  \caption{Scaling training data 5-way 1-shot results for models trained on COCO. Datasets marked on the left of each plot are the evaluation dataset. Last plot is the average performance. X-axis is the percentage of the total training data and y-axis is the 5-way 1-shot accuracy.}
  \label{fig:scaling_training_data_coco_5way_1shot}
\end{figure}

\clearpage




\clearpage

\begin{figure}
  \begin{center}
  \includegraphics[width=0.87\linewidth]{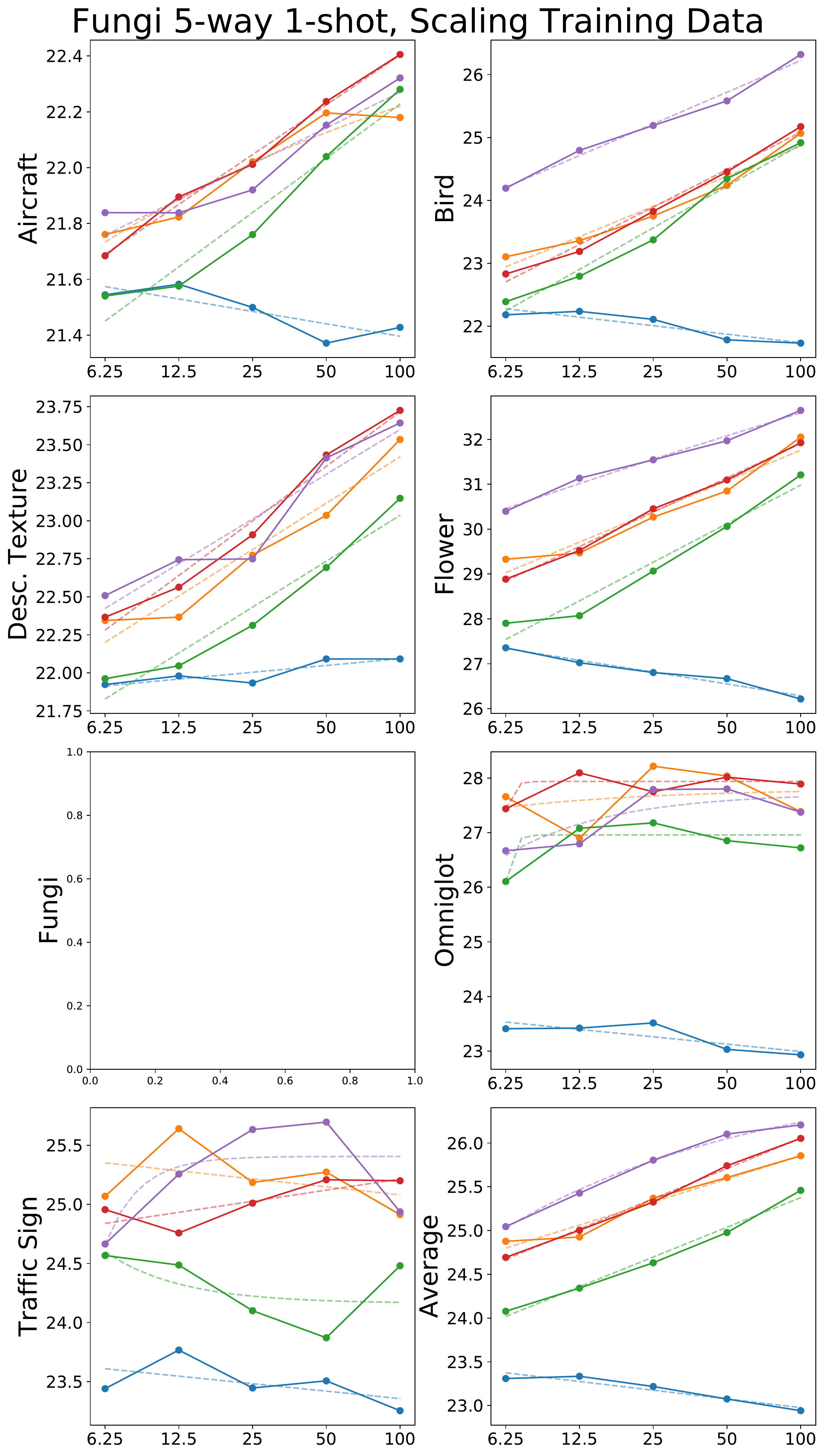}
  \includegraphics[width=0.55\linewidth]{figure/natural_images_scaling_amount_of_training_data_5way_5shot_legend.pdf}
  \end{center}
  \caption{Scaling training data 5-way 1-shot results for models trained on Fungi. Datasets marked on the left of each plot are the evaluation dataset. Last plot is the average performance. X-axis is the percentage of the total training data and y-axis is the 5-way 1-shot accuracy.}
  \label{fig:scaling_training_data_fungi_5way_1shot}
\end{figure}

\clearpage

\begin{figure}
  \begin{center}
  \includegraphics[width=0.87\linewidth]{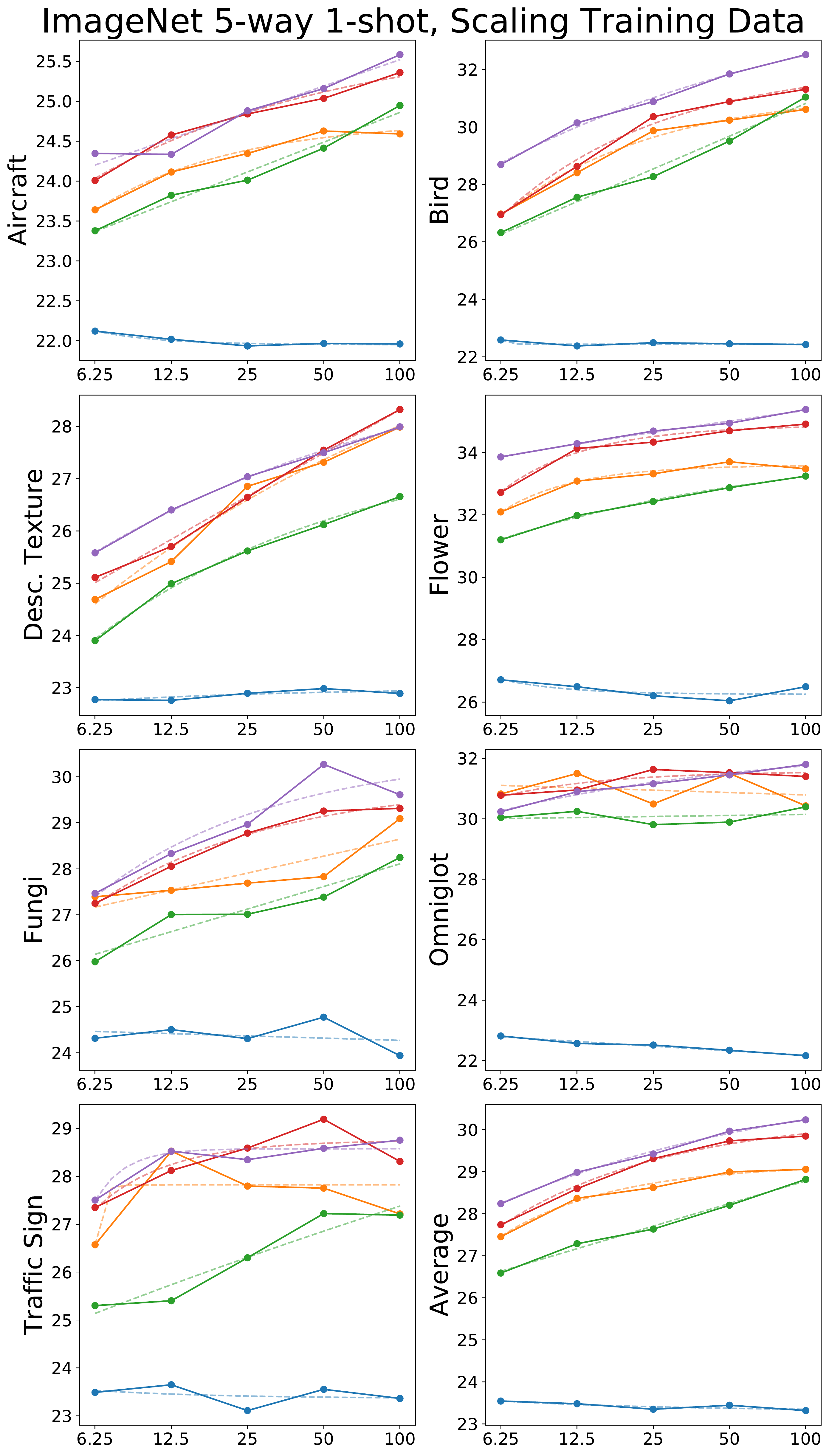}
  \includegraphics[width=0.55\linewidth]{figure/natural_images_scaling_amount_of_training_data_5way_5shot_legend.pdf}
  \end{center}
  \caption{Scaling training data 5-way 1-shot results for models trained on ImageNet. Datasets marked on the left of each plot are the evaluation dataset. Last plot is the average performance. X-axis is the percentage of the total training data and y-axis is the 5-way 1-shot accuracy.}
  \label{fig:scaling_training_data_imagenet_5way_1shot}
\end{figure}

\clearpage


\clearpage

\begin{figure}
  \begin{center}
  \includegraphics[width=0.86\linewidth]{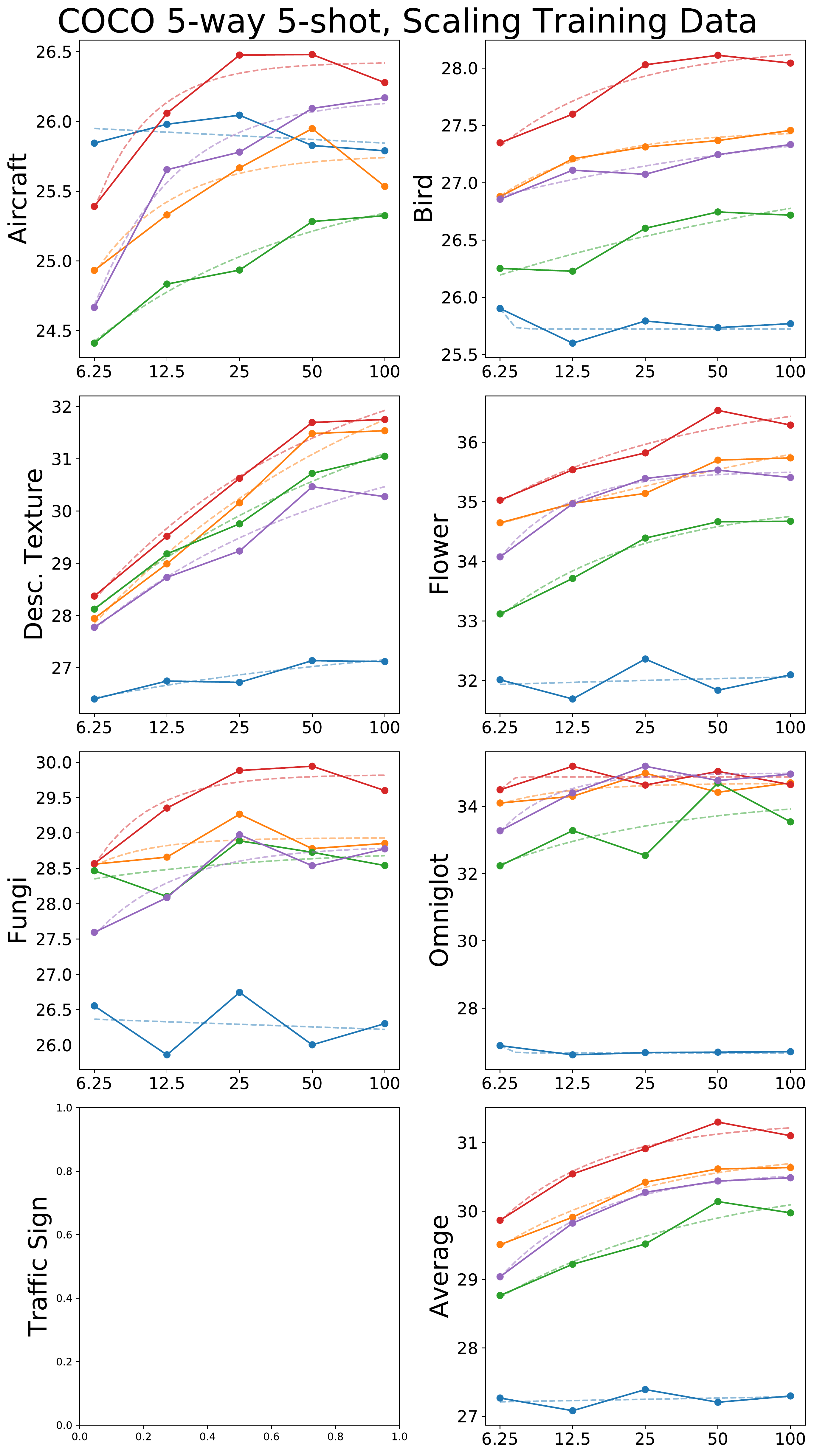}
  \includegraphics[width=0.55\linewidth]{figure/natural_images_scaling_amount_of_training_data_5way_5shot_legend.pdf}
  \end{center}
  \caption{Scaling training data 5-way 5-shot results for models trained on COCO. Datasets marked on the left of each plot are the evaluation dataset. Last plot is the average performance. X-axis is the percentage of the total training data and y-axis is the 5-way 5-shot accuracy.}
  \label{fig:scaling_training_data_coco_5way_5shot}
\end{figure}

\clearpage





\begin{figure}
  \begin{center}
  \includegraphics[width=0.87\linewidth]{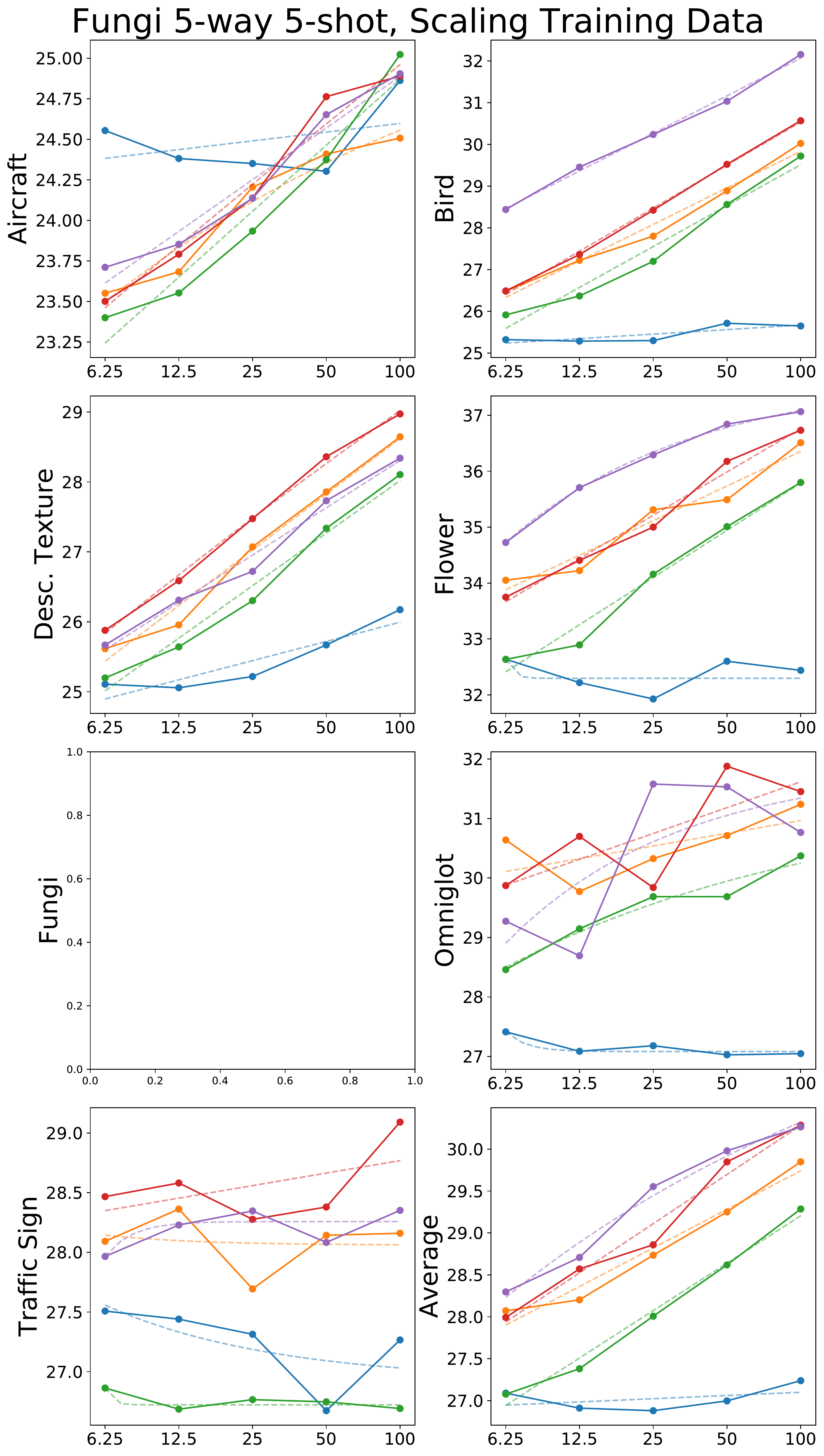}
  \includegraphics[width=0.55\linewidth]{figure/natural_images_scaling_amount_of_training_data_5way_5shot_legend.pdf}
  \end{center}
  \caption{Scaling training data 5-way 5-shot results for models trained on Fungi. Datasets marked on the left of each plot are the evaluation dataset. Last plot is the average performance. X-axis is the percentage of the total training data and y-axis is the 5-way 5-shot accuracy.}
  \label{fig:scaling_training_data_fungi_5way_5shot}
\end{figure}

\clearpage

\begin{figure}
  \begin{center}
  \includegraphics[width=0.86\linewidth]{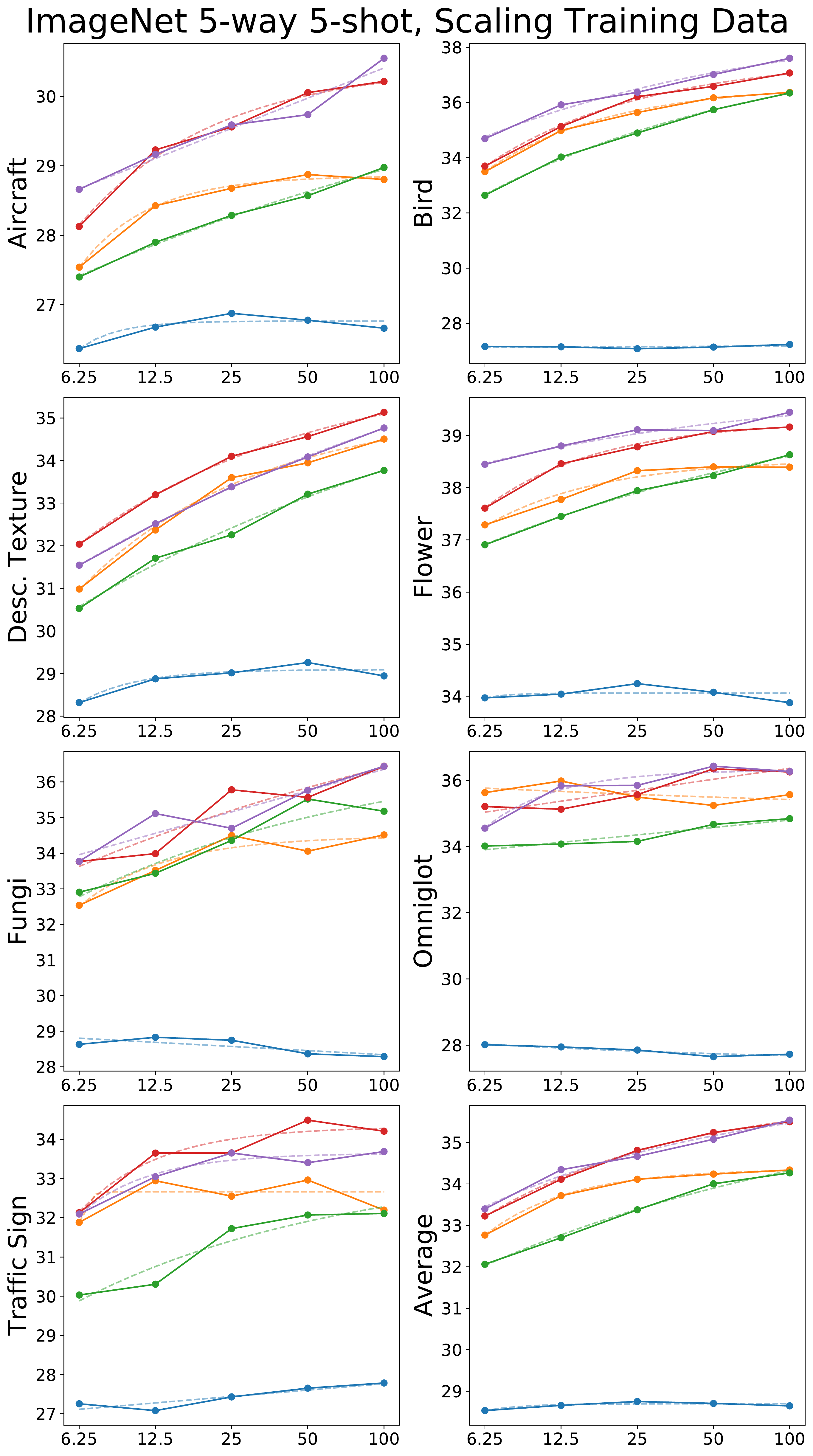}
  \includegraphics[width=0.55\linewidth]{figure/natural_images_scaling_amount_of_training_data_5way_5shot_legend.pdf}
  \end{center}
  \caption{Scaling training data 5-way 5-shot results for models trained on ImageNet. Datasets marked on the left of each plot are the evaluation dataset. Last plot is the average performance. X-axis is the percentage of the total training data and y-axis is the 5-way 5-shot accuracy.}
  \label{fig:scaling_training_data_imagenet_5way_5shot}
\end{figure}

\clearpage

\begin{figure}
  \begin{center}
  \includegraphics[width=0.86\linewidth]{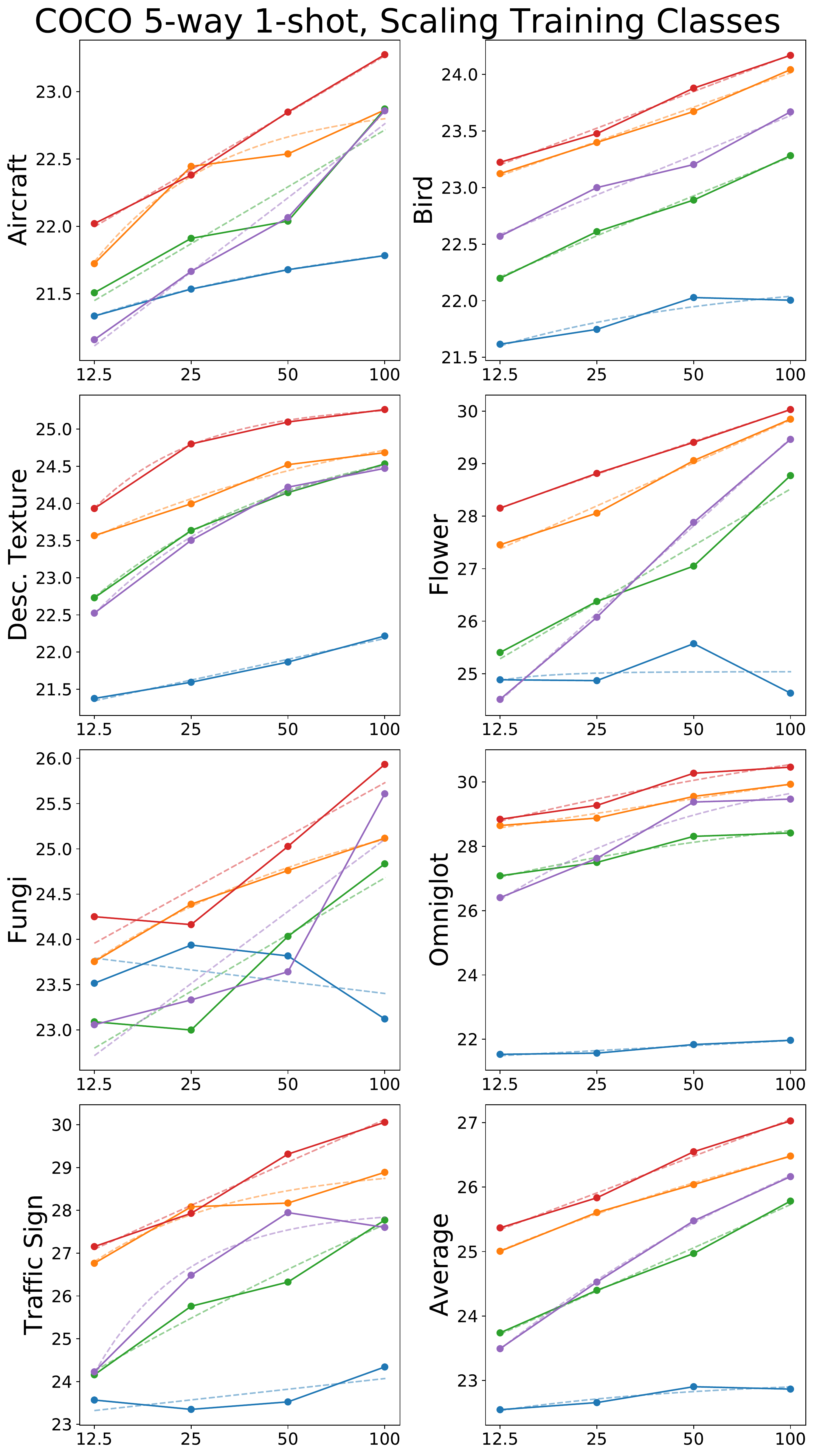}
  \includegraphics[width=0.55\linewidth]{figure/natural_images_scaling_amount_of_training_data_5way_5shot_legend.pdf}
  \end{center}
  \caption{Scaling training classes 5-way 1-shot results for models trained on COCO. Datasets marked on the left of each plot are the evaluation dataset. Last plot is the average performance. X-axis is the percentage of the total training classes and y-axis is the 5-way 1-shot accuracy.}
  \label{fig:scaling_training_classes_coco_5way_1shot}
\end{figure}

\clearpage





\begin{figure}
  \begin{center}
  \includegraphics[width=0.86\linewidth]{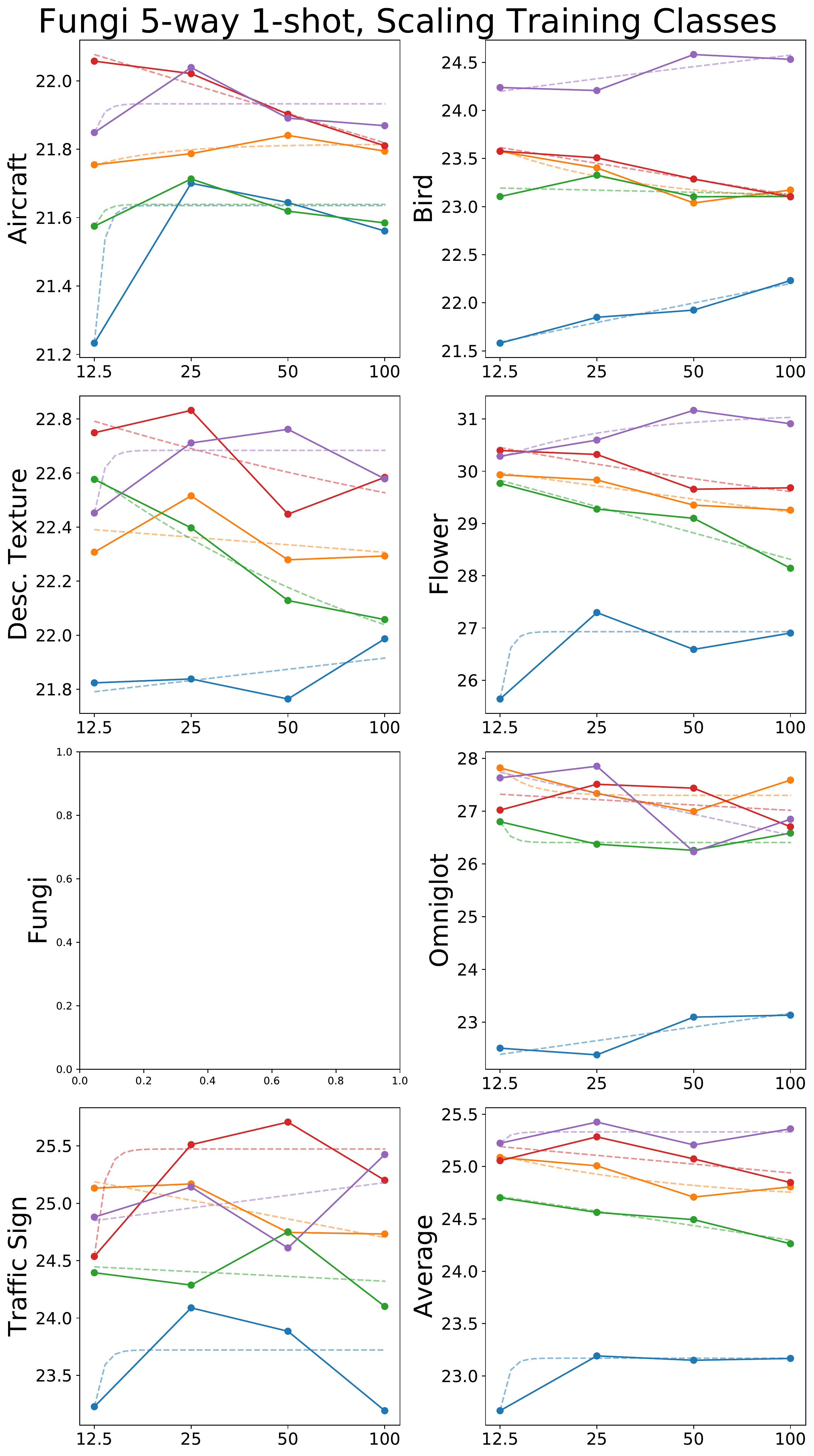}
  \includegraphics[width=0.55\linewidth]{figure/natural_images_scaling_amount_of_training_data_5way_5shot_legend.pdf}
  \end{center}
  \caption{Scaling training classes 5-way 1-shot results for models trained on Fungi. Datasets marked on the left of each plot are the evaluation dataset. Last plot is the average performance. X-axis is the percentage of the total training classes and y-axis is the 5-way 1-shot accuracy.}
  \label{fig:scaling_training_classes_fungi_5way_1shot}
\end{figure}

\clearpage

\begin{figure}
  \begin{center}
  \includegraphics[width=0.87\linewidth]{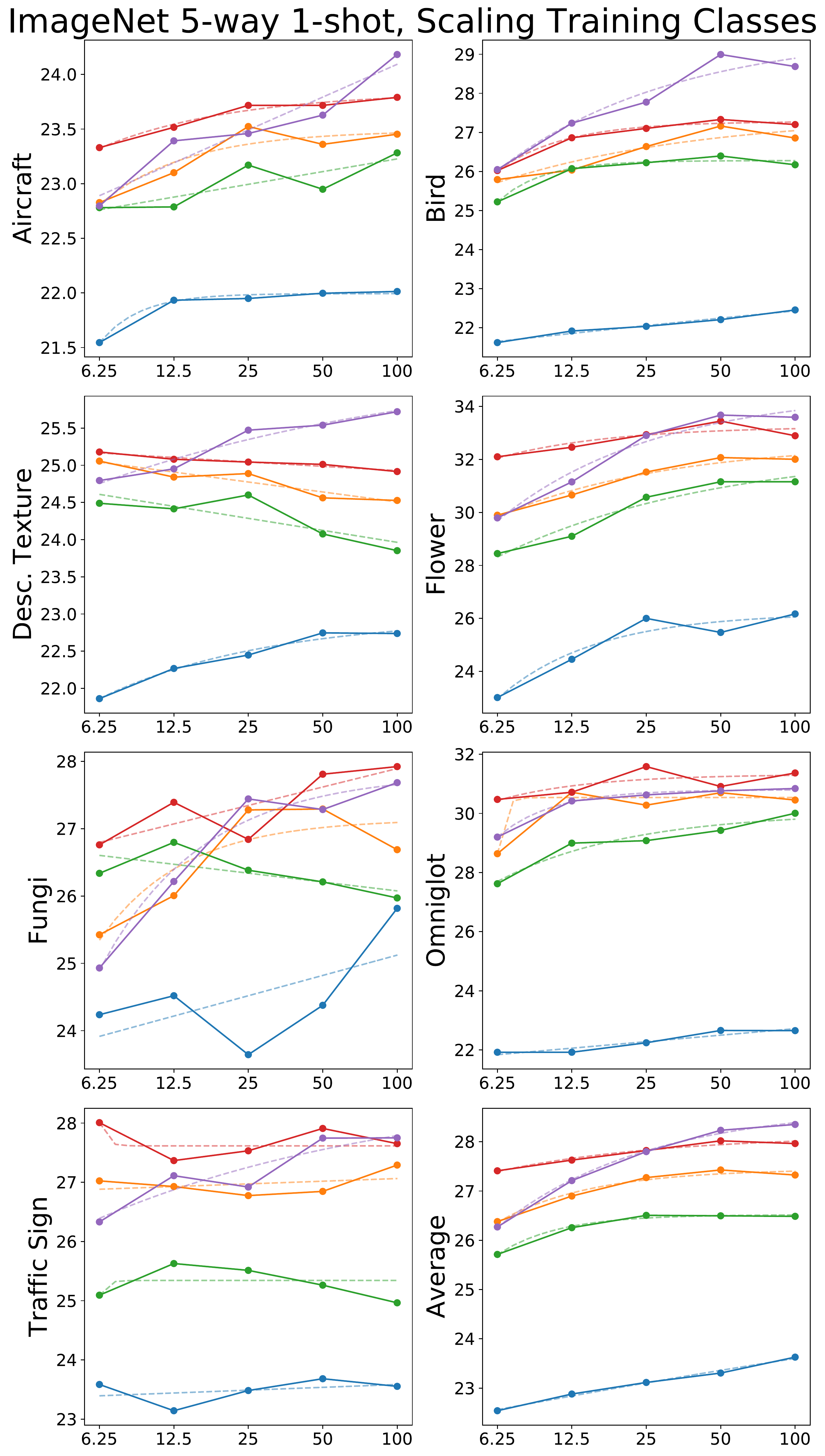}
  \includegraphics[width=0.55\linewidth]{figure/natural_images_scaling_amount_of_training_data_5way_5shot_legend.pdf}
  \end{center}
  \caption{Scaling training classes 5-way 1-shot results for models trained on ImageNet. Datasets marked on the left of each plot are the evaluation dataset. Last plot is the average performance. X-axis is the percentage of the total training classes and y-axis is the 5-way 1-shot accuracy.}
  \label{fig:scaling_training_classes_imagenet_5way_1shot}
\end{figure}

\clearpage

\begin{figure}
  \begin{center}
  \includegraphics[width=0.87\linewidth]{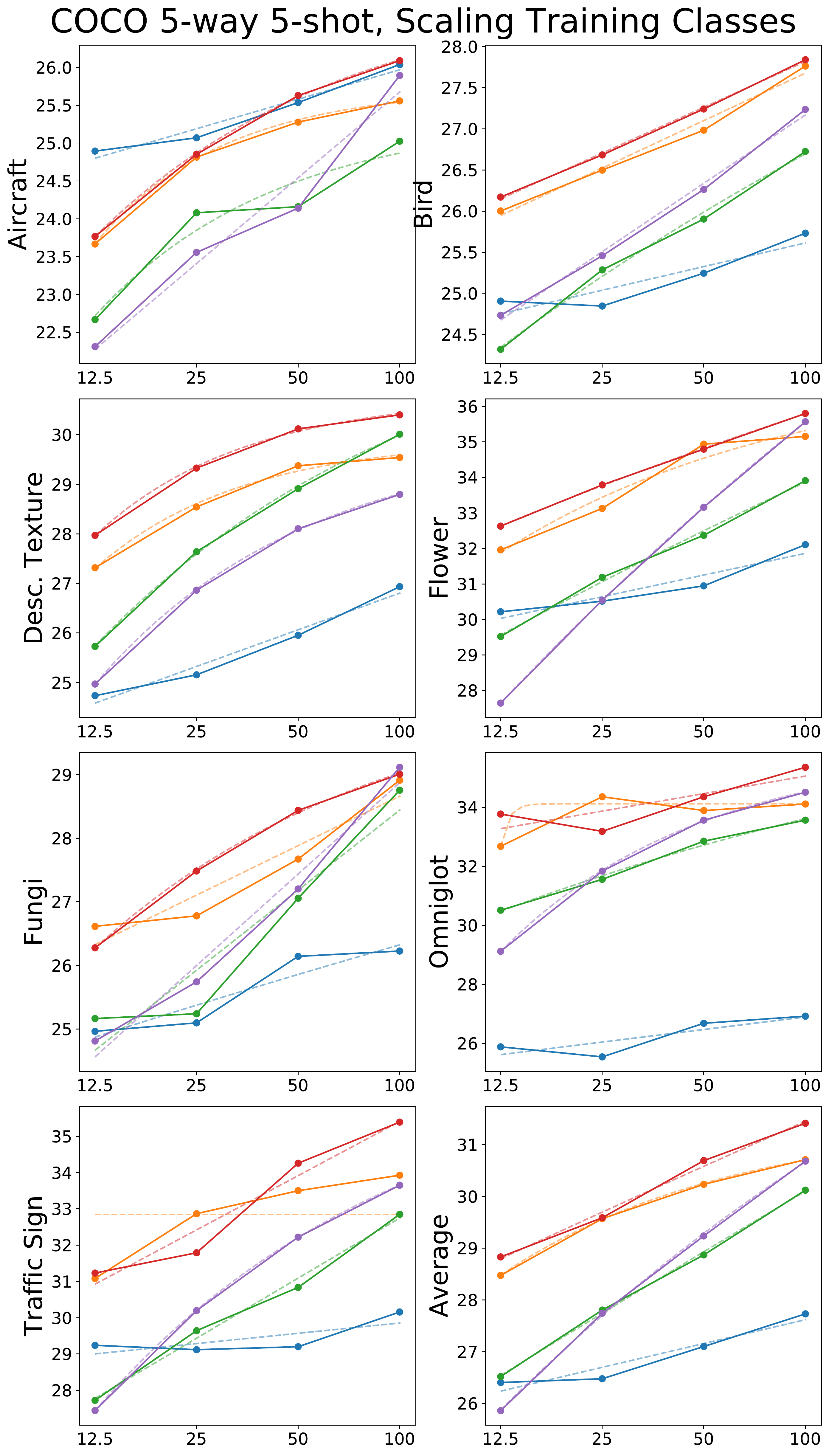}
  \includegraphics[width=0.55\linewidth]{figure/natural_images_scaling_amount_of_training_data_5way_5shot_legend.pdf}
  \end{center}
  \caption{Scaling training classes 5-way 5-shot results for models trained on COCO. Datasets marked on the left of each plot are the evaluation dataset. Last plot is the average performance. X-axis is the percentage of the total training classes and y-axis is the 5-way 5-shot accuracy.}
  \label{fig:scaling_training_classes_coco_5way_5shot}
\end{figure}

\clearpage





\begin{figure}
  \begin{center}
  \includegraphics[width=0.86\linewidth]{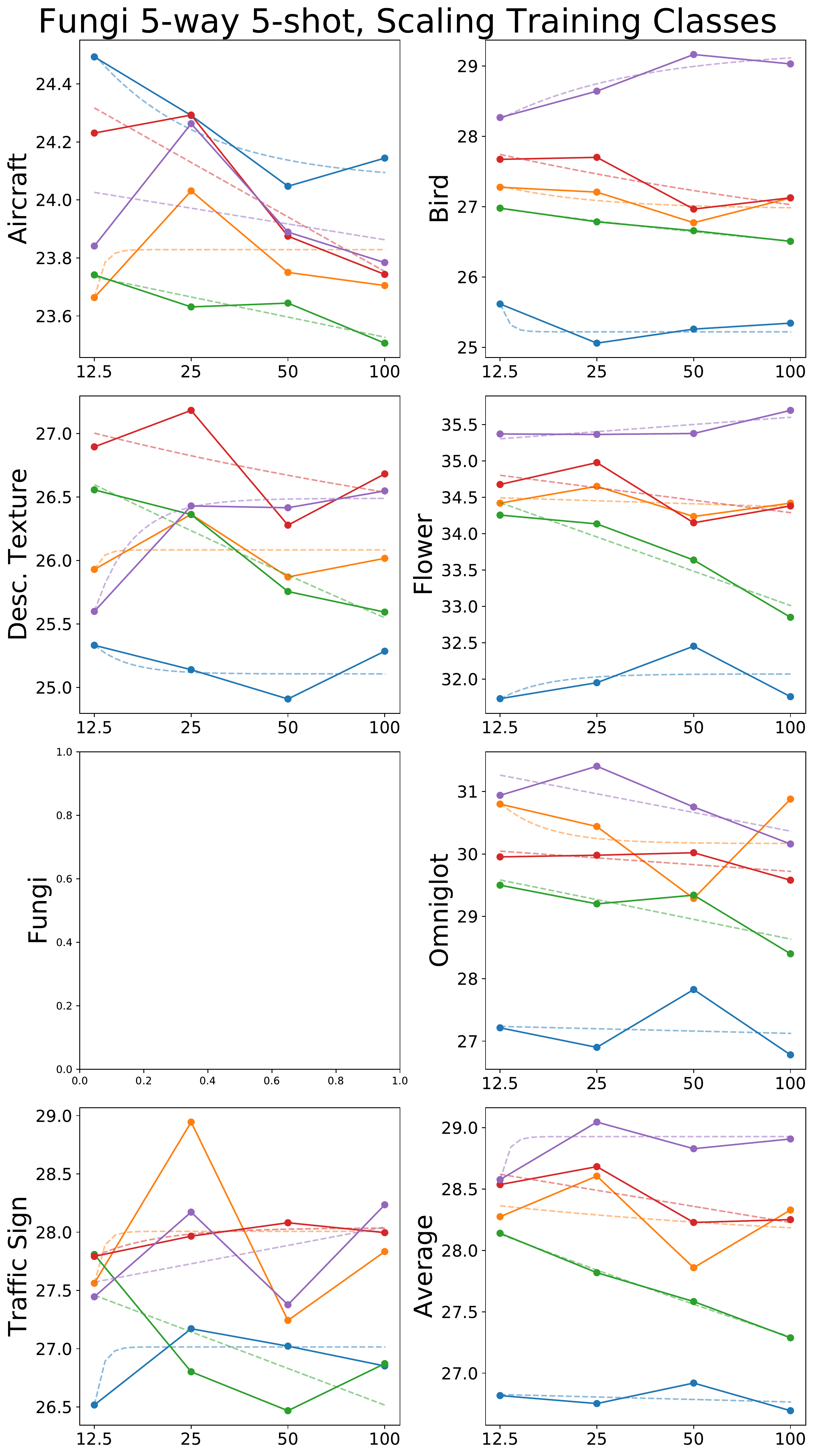}
  \includegraphics[width=0.55\linewidth]{figure/natural_images_scaling_amount_of_training_data_5way_5shot_legend.pdf}
  \end{center}
  \caption{Scaling training classes 5-way 5-shot results for models trained on Fungi. Datasets marked on the left of each plot are the evaluation dataset. Last plot is the average performance. X-axis is the percentage of the total training classes and y-axis is the 5-way 5-shot accuracy.}
  \label{fig:scaling_training_classes_fungi_5way_5shot}
\end{figure}

\clearpage

\begin{figure}
  \begin{center}
  \includegraphics[width=0.87\linewidth]{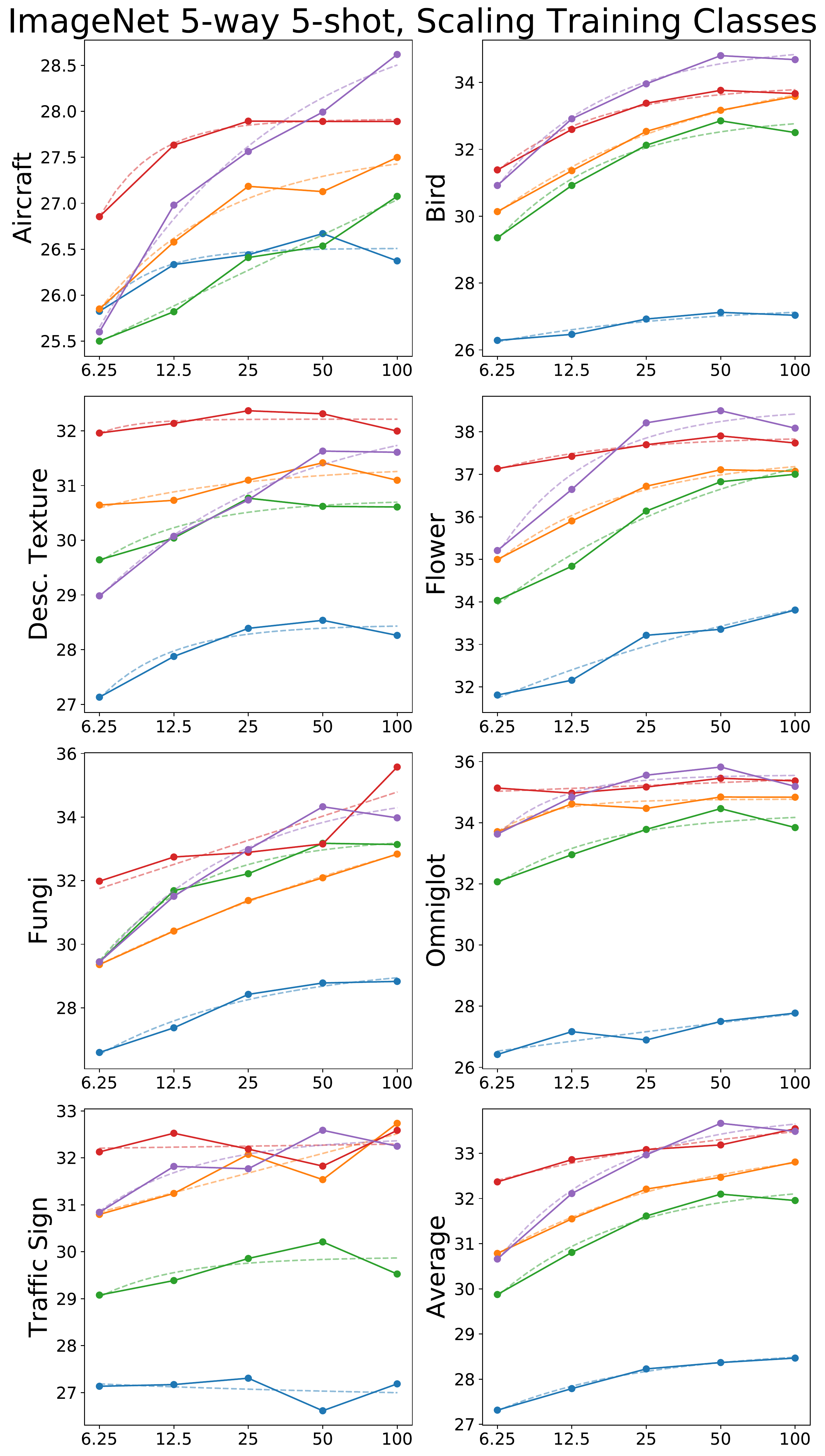}
  \includegraphics[width=0.55\linewidth]{figure/natural_images_scaling_amount_of_training_data_5way_5shot_legend.pdf}
  \end{center}
  \caption{Scaling training classes 5-way 5-shot results for models trained on ImageNet. Datasets marked on the left of each plot are the evaluation dataset. Last plot is the average performance. X-axis is the percentage of the total training classes and y-axis is the 5-way 5-shot accuracy.}
  \label{fig:scaling_training_classes_imagenet_5way_5shot}
\end{figure}

\clearpage

\end{document}